\documentclass{WileyMSP-template}
\usepackage{float}
\usepackage{hyperref}
\usepackage{booktabs}
\usepackage{xcolor}
\usepackage{multirow}
\usepackage{makecell}
\usepackage{graphicx}
\usepackage{amssymb}
\usepackage{tabularx}
\usepackage{array}
\usepackage[table]{xcolor}

\begin{document}

\pagestyle{fancy}
\rhead{\includegraphics[width=2.5cm]{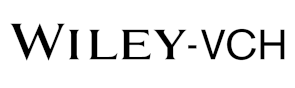}}

\title{TacVerse: A Multi-Sensor Dataset and Benchmark for\\Cross-Sensor Vision-Based Tactile Perception}
\maketitle


\author{Lan Wei,}
\author{Gurmeher Khurana,}
\author{Sirine Bhouri,}
\author{Wenhao Hong,}
\author{Zeyuan Xin,}
\author{Qingzheng Cong,}
\author{Wen Fan,}
\author{Yanzheng Xiang,}
\author{Dandan Zhang*}

\begin{affiliations}
L. Wei, G. Khurana, S. Bhouri, W. Hong, Z. Xin, W. Fan and Dr D. Zhang\\
Imperial College London, UK. \\
Email Address: d.zhang17@imperial.ac.uk\\

Q. Cong\\
Queen Mary University of London, UK\\

Y. Xiang\\
King's College London, UK\\
\end{affiliations}


\keywords{Vision-Based Tactile Sensing, Multi-Sensor Tactile Dataset, Cross-Sensor Transfer, Self-Supervised Learning}


\begin{abstract}
Vision-based tactile sensors (VBTSs) enable robots to infer contact geometry and force-related cues by imaging deformation through an internal camera, yet generalisation across sensor designs remains poorly understood. We present TacVerse, a multi-sensor dataset and benchmark for cross-sensor vision-based tactile perception. 
The dataset contains 106,800 tactile images from seven VBTSs and supports three downstream tasks: shape classification, grating classification, and force regression. 
Experiments are conducted under three settings: within-sensor training, zero-shot cross-sensor transfer, and few-shot adaptation. 
Strong within-sensor performance across all tasks indicates that the collected tactile observations are informative for the target objectives. 
Direct cross-sensor transfer, however, leads to substantial degradation. Shape classification is comparatively robust, whereas grating classification and force regression are more sensitive to sensor shift. Few-shot adaptation for force regression consistently improves performance on unseen target sensors but does not fully close the gap to within-sensor upper bounds.
A representation study further shows that MAE (Masked Autoencoder) pretraining provides the most consistent gains across tasks and sensors. TacVerse provides a controlled testbed for studying sensor shift, data-efficient adaptation, and self-supervised learning in tactile perception.
\end{abstract}

\section{Introduction}
Tactile sensing is essential for robots to interact reliably with the physical world~\cite{dahiya2012robotic}. While vision provides rich global context, it often cannot directly recover local contact state, surface deformation, incipient slip, or contact forces once interaction occurs at the robot--object interface~\cite{zhang2025safe}. 
VBTSs address this limitation by turning touch into an imaging problem: an embedded camera observes a deformable skin under controlled illumination and converts contact into high-dimensional visual observations 
that can be interpreted using computer vision, physical modeling, computational simulation, and machine learning methods, such as data-driven deep learning~\cite{zhang2022hardware}.
This mechanism has made VBTSs an increasingly important sensing modality for fine-grained perception and contact-rich manipulation.

A central obstacle to progress, however, is sensor heterogeneity. 
Although VBTSs share the common principle of imaging contact-induced deformation, they differ substantially in their contact-to-image transduction mechanisms, which determine the physical cues they are most sensitive to and how these cues are encoded as image features~\cite{li2025classification}.
Following a modality-driven taxonomy, vision-based tactile sensing can be broadly grouped into three sensing principles: \emph{Intensity Mapping Method} (IMM), which infers shape or pressure from reflected-light variations; \emph{Marker Displacement Method} (MDM), which estimates deformation by tracking printed or embedded markers; and 
\emph{Modality Fusion Method} (MFM), which uses transparent ``see-through'' skins and tailored illumination to expose the contact interface, fusing visual appearance with tactile cues and, in some designs, complementary modalities such as proximity and temperature sensing~\cite{fan2025crystaltac}.
In practice, many sensors combine these principles in hybrid designs~\cite{li2025classification}. As a result, sensors can differ markedly in optical design, illumination, gel structure, marker configuration, spatial resolution, and contact geometry~\cite{xin2025vision}, which in turn creates substantial appearance and distribution shifts across datasets~\cite{albini2025representing}. Consequently, many existing tactile datasets are effectively sensor-specific, making it difficult to determine whether learned models capture transferable contact information or merely overfit to sensor-specific artifacts~\cite{li2025object}. The challenge is further amplified by the cost of tactile data collection, which requires repeated physical interaction, scales slowly, and may introduce sensor wear over time.

\begin{figure}[!t]
\centering
\includegraphics[width=0.9\columnwidth]{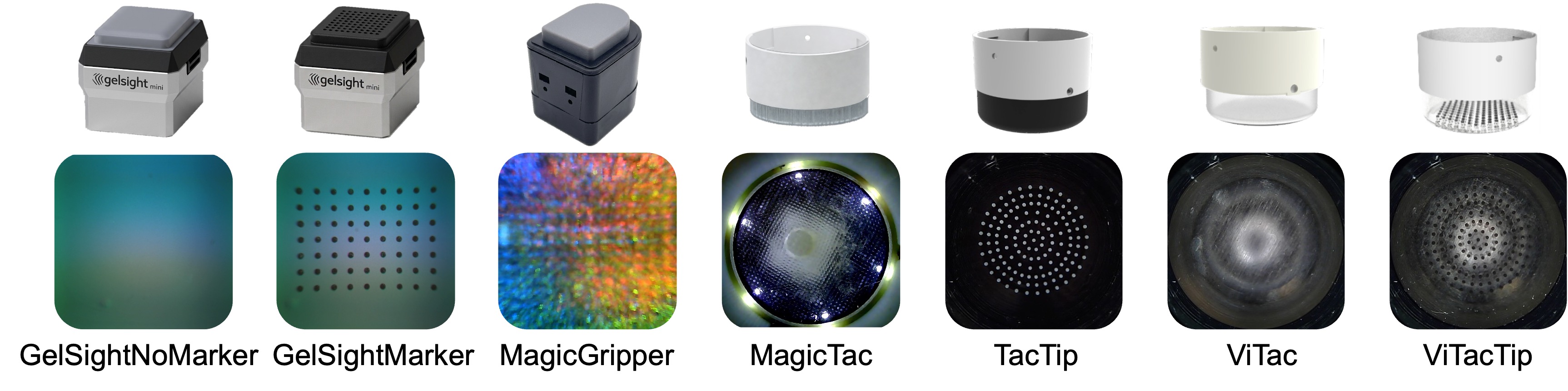}
\vspace{-10pt}
\caption{Overview of the seven vision-based tactile sensors in TacVerse and representative tactile images captured by each sensor. 
The suite covers diverse sensing principles and hybrid configurations, creating substantial cross-sensor variation that motivates our benchmark for tactile perception and transfer.}
\label{fig1}
\end{figure}

A second challenge is evaluation. Recent work has substantially advanced tactile representation learning and transfer, but existing resources often focus on different aspects of the problem. T3 studies transfer across heterogeneous sensor--task pairs at large scale~\cite{zhao2024transferable}; AnyTouch emphasises aligned multi-sensor data and unified representation learning~\cite{feng2025anytouch}; Sparsh shows that self-supervised pretraining can significantly improve tactile perception~\cite{higuera2024sparsh}; and SITR highlights the importance of sensor-invariant transfer across tactile domains~\cite{gupta2025sensor}. These efforts are important, but they do not fully answer a compact and practically relevant set of questions under one controlled setup: How well is a task learned within a sensor? How much performance is lost under sensor shift? How much can limited target supervision recover? And can one self-supervised tactile representation support multiple downstream tactile-vision tasks?

In this paper, we present TacVerse, a multi-sensor dataset and benchmark for cross-sensor vision-based tactile perception. As shown in Fig.~\ref{fig1}, TacVerse contains 106,800 tactile images collected from seven VBTS platforms spanning multiple sensing families and hybrid configurations, including GelSightNoMarker (IMM), GelSightMarker (IMM+MDM)~\cite{yuan2017GelSight}, MagicGripper (IMM+MDM+MFM)~\cite{fan2025magicgripper}, MagicTac (IMM+MDM+MFM)~\cite{fan2024magictac}, TacTip (MDM)~\cite{lepora2021soft}, ViTac (IMM+MFM)~\cite{luo2021vitac}, and ViTacTip (MDM+MFM)~\cite{fan2024vitactip}. 
The dataset is organised into three downstream tasks: 9-way shape classification, 30-way grating classification, and force regression. 
Rather than proposing another sensor-specific architecture, TacVerse is designed to make cross-sensor tactile learning measurable, reproducible, and comparable.

We study three complementary settings: within-sensor training, zero-shot cross-sensor transfer, and few-shot adaptation. Together, these settings separate task learnability from sensor transfer while keeping the benchmark compact and easy to reproduce. Across all three tasks, strong within-sensor results show that the collected tactile observations are informative for the target objectives. By contrast, direct cross-sensor transfer causes substantial performance degradation, revealing a clear sensor-shift gap. Among the three tasks, shape classification is comparatively robust, whereas grating classification and force regression are more sensitive to sensor mismatch. Few-shot adaptation for force regression substantially improves performance on unseen target sensors, but it does not fully close the gap to the within-sensor upper bound. A separate representation study further shows that MAE pretraining provides the most consistent gains across tasks and sensors, suggesting that self-supervised tactile pretraining offers a strong unified initialisation.


Our contributions are fourfold:
\begin{enumerate}
    \item We present TacVerse, a multi-sensor tactile dataset with 106,800 images collected from seven VBTS platforms spanning multiple sensing families and hybrid sensing principles.
    \item We organise TacVerse into three benchmark tasks: 9-way shape classification, 30-way grating classification, and force regression, to support evaluation across both classification and regression settings.
    \item We establish a unified benchmark for cross-sensor tactile perception with three complementary settings: within-sensor training, zero-shot cross-sensor transfer, and few-shot adaptation.
    \item We provide an empirical study showing that within-sensor learning is strong, direct cross-sensor transfer remains challenging, limited target supervision substantially improves adaptation, and MAE pretraining yields the most consistent representation across tasks and sensors.
\end{enumerate}

\section{Related Work}
\begin{table*}[t]
    \centering
    
    \caption{Comparison of representative recent multi-sensor vision-based tactile datasets and benchmarks in terms of sensor coverage, data scale, task focus, and primary objective.}
    \label{tab:dataset_comparison}
    \small
    \setlength{\tabcolsep}{4pt}
    \begin{tabularx}{\textwidth}{
        @{}l
        c
        c
        >{\raggedright\arraybackslash}p{3cm}
        >{\raggedright\arraybackslash}X
    }
        \toprule
        Resource &
        Sensors &
        Scale &
        Task focus &
        Dataset objective  \\
        \midrule

        FoTa~\cite{zhao2024transferable} &
        13 &
        3,083,452 &
        Recognition and pose estimation &
        Designed for large-scale multi-sensor and multi-task learning of transferable tactile representations.  \\
        \midrule
        
        TacQuad~\cite{feng2025anytouch} &
        4 &
        72,606 &
        Property recognition and interaction prediction &
        Designed to learn unified touch-vision-language representations from aligned multi-sensor and multimodal observations.\\
        \midrule        

        TacBench~\cite{higuera2024sparsh} &
        3&
        460,000 &
        Tactile perception and manipulation &
        Designed to evaluate self-supervised tactile representations across multiple sensors and downstream tasks.\\
        \midrule        

        \textbf{TacVerse (ours)} &
        \textbf{7} &
        \textbf{106,800} &
        \textbf{Shape, grating, and force perception} &
        \textbf{Designed to quantify sensor shift, direct cross-sensor transfer, and few-shot adaptation under controlled task settings.} \\
        \bottomrule
    \end{tabularx} 
\end{table*}

\subsection{Vision-based tactile datasets and benchmarks}

Prior work in tactile perception has relied on sensor- and task-specific datasets, making it difficult to compare models across tactile platforms or systematically study representation transfer under sensor shift. 
Recent multi-sensor resources have expanded the scale and scope of tactile learning in several complementary directions, as summarised in Table~\ref{tab:dataset_comparison}.
T3 introduces FoTa, a unified-format dataset containing more than 3 million data points from 13 sensors and 11 tasks. It focuses on recognition and pose-estimation tasks and is designed for large-scale multi-sensor and multi-task learning of transferable tactile representations~\cite{zhao2024transferable}. 
AnyTouch introduces TacQuad, an aligned multimodal dataset collected using four visuo-tactile sensors. It focuses on tactile-property recognition and interaction prediction, with the objective of learning unified touch-vision-language representations from aligned multi-sensor observations~\cite{feng2025anytouch}. 
Sparsh introduces TacBench, a six-task benchmark constructed from more than 460,000 tactile images. It focuses on tactile perception and manipulation tasks and is designed to evaluate self-supervised tactile representations across sensors and downstream applications~\cite{higuera2024sparsh}.
Other benchmark efforts address related but distinct objectives. TacEva evaluates the hardware-level performance of vision-based tactile sensors~\cite{cong2026taceva}, Tactile MNIST focuses on active tactile perception~\cite{schneider2025tactile}, and ManiFeel evaluates visuotactile manipulation policies in simulation with real-world validation~\cite{luu2025manifeel}.
In contrast, TacVerse provides task-aligned observations across multiple sensors, with shared task definitions and label spaces within each task to isolate sensor variation from task or annotation mismatch.

\subsection{Tactile representation learning}
Beyond supervised tactile perception, a growing line of work studies transferable tactile representations. 
UniTouch learns unified multimodal tactile embeddings by aligning touch to pretrained image representations that already connect to other modalities, and uses sensor-specific tokens to accommodate heterogeneous tactile sensors~\cite{yang2024binding}. 
MViTac studies self-supervised contrastive pretraining for joint visual--tactile representation learning~\cite{dave2024multimodal}. 
Contrastive Touch-to-Touch Pretraining (CTTP) aligns paired observations from GelSlim and Soft Bubble in a shared embedding space through contrastive learning, supporting downstream pose estimation and classification as well as deployment across the two sensors without additional training~\cite{rodriguez2025contrastive}.
UniT learns a compact tactile latent space with a VQ-based generative model and shows that representations learned from simple tactile data can transfer to downstream perception and manipulation tasks~\cite{xu2025unit}. 
Sparsh provides the most systematic recent evaluation of self-supervised learning for vision-based tactile sensing, showing that tactile pretraining can significantly improve downstream performance and that latent-space objectives such as DINO and I-JEPA are especially effective~\cite{higuera2024sparsh}. 
Together, these works establish the value of tactile pretraining, but they primarily develop representation-learning objectives or evaluate selected sensor pairs and downstream tasks rather than providing a controlled multi-sensor benchmark across several task-aligned settings.

\begin{figure}[!t]
\centering
\includegraphics[width=1\columnwidth]{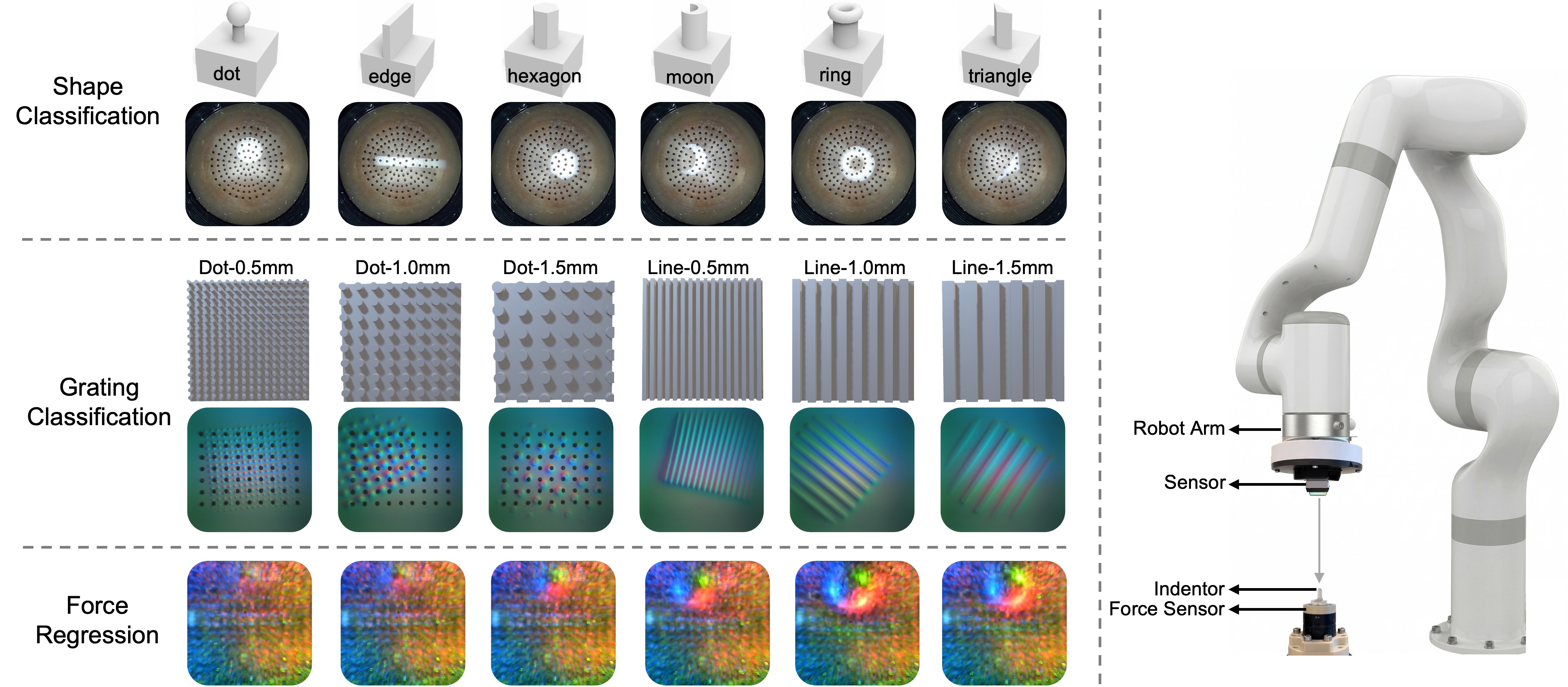}
\caption{Representative samples from the three TacVerse benchmark tasks. 
Top: Shape classification on ViTacTip, showing six example contact shapes and their corresponding tactile observations. 
Middle: Grating classification on GelSightNoMarker and GelSightMarker, illustrating representative dot and line gratings with different spatial scales together with tactile images. 
Bottom and right: Force regression on MagicGripper, showing example tactile observations under different contact-force conditions. 
These tasks span both discrete recognition and continuous estimation, providing complementary testbeds for cross-sensor tactile perception.}
\label{fig2}
\end{figure}

\subsection{Cross-sensor transfer and sensor-agnostic tactile learning}

A more closely related line of work directly addresses transfer across heterogeneous tactile sensors. 
T3 learns across multi-sensor and multi-task data using sensor-specific encoders, a shared transformer trunk, and task-specific decoders, demonstrating that tactile representations can scale across diverse sensor--task combinations~\cite{zhao2024transferable}. 
AnyTouch further studies unified static--dynamic multi-sensor representations by combining masked modelling, multimodal alignment, and cross-sensor matching on aligned data from multiple visuo-tactile sensors~\cite{feng2025anytouch}. 
SITR moves one step closer to explicit sensor invariance: it trains a transformer on simulated sensor variations and uses minimal calibration to achieve zero-shot transfer across optical tactile sensors~\cite{gupta2025sensor}. 
Touch-to-Touch Translation learns an explicit mapping from DIGIT images to CySkin outputs under corresponding physical stimuli, demonstrating signal-level translation between camera-based and capacitive tactile technologies~\cite{grella2025touch}. 
GenForce unifies heterogeneous tactile signals into shared marker representations and uses marker-to-marker translation to transfer force-prediction capabilities from calibrated sensors to homogeneous and heterogeneous target sensors without exhaustive target force-data collection~\cite{chen2026training}.
These works demonstrate cross-sensor learning through shared representations, simulated sensor invariance, signal translation, and force-specific transfer, but their primary contributions lie in transfer-model design for selected sensor configurations and tasks. 
TacVerse complements these model-centric approaches with an architecture-controlled benchmark that measures sensor-shift degradation through within-sensor and zero-shot evaluation, as well as recovery from limited target supervision through few-shot force-regression adaptation.

\section{Dataset}
TacVerse is a multi-sensor dataset for cross-sensor vision-based tactile perception. It contains 106,800 labelled tactile images collected from seven vision-based tactile sensors: 
GelSightNoMarker (GelSight Mini without markers) (16,917 images), GelSightMarker (GelSight Mini with markers) (15,487 images), MagicGripper (16,892 images), MagicTac (12,496 images), TacTip (11,000 images), ViTac (11,000 images), and ViTacTip (23,008 images). These sensors span multiple sensing principles and hybrid configurations, creating substantial variation in tactile appearance, deformation patterns, and force-related cues across platforms. Representative examples are shown in Fig.~\ref{fig2}.
To support controlled comparison across sensing platforms, TacVerse is organised into three downstream tasks: shape classification (30,094 images), grating classification (40,509 images), and force regression (36,197 images). This task-based organisation keeps the data collection procedure consistent within each benchmark while preserving meaningful sensor diversity for cross-sensor evaluation.

\textbf{Shape classification:}
The shape subset was collected by pressing a set of 3D-printed indenters with diverse 3D geometries against the tactile sensors~\cite{jianu2022reducing}. For each interaction, the robot recorded the tactile image generated at contact. The resulting data preserve the contact location and local object contour and form a 9-way shape-classification dataset over the shape categories shared by all sensors.

\textbf{Grating classification:}
The grating subset was collected using 3D-printed samples with fine spatial patterns, including line- and dot-like gratings. The grating widths and spacings range from 0.3 mm to 1.75 mm in 0.05 mm increments. During data collection, the sensor was mounted on the robot end-effector and pressed against each sample 100 times while varying the tool-centre-point yaw orientation. 
The resulting dataset captures variation in both spatial frequency and contact orientation across two grating configurations, which are line-like and dot-like, each comprising 30 grating classes.

\textbf{Force regression:}
The force subset was collected using a six-axis force/torque (F/T) transducer (M3813B) as the ground-truth measurement device. A 3D-printed spherical indenter with a 2 mm radius was rigidly attached to the transducer, while the tactile sensor was mounted on a robot arm. The robot approached the indenter along the vertical axis in steps of 0.1 mm until contact was established, then applied additional small displacements in the $x$-$y$ plane to vary contact direction and capture both normal and shear responses. Each tactile image was synchronised with the transducer readings, and the measured force components $(F_x, F_y, F_z)$ serve as the regression targets.

\section{Experiment}
\subsection{Experiment Setting}
\begin{figure}[!t]
\centering
\includegraphics[width=0.9\columnwidth]{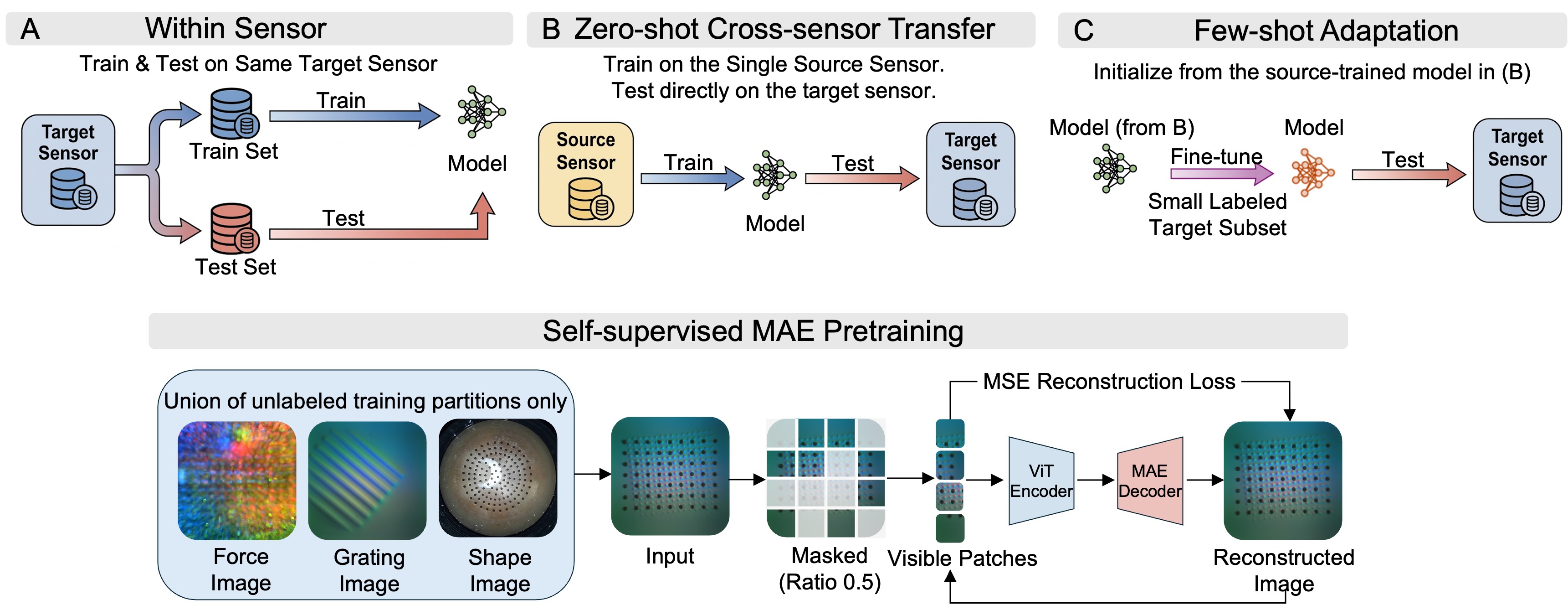}
\caption{TacVerse benchmark protocols and self-supervised pretraining setup. The figure illustrates three evaluation protocols: (A) within-sensor training, (B) zero-shot cross-sensor transfer, and (C) few-shot adaptation on the target sensor. It also shows the MAE-based pretraining pipeline, in which a shared tactile encoder is pretrained on the combined images from the three tasks and then transferred across downstream tactile-perception tasks.}
\label{fig3}
\end{figure}

Figure~\ref{fig3} summarises the benchmark setup. The experiments are divided into two parts: a transfer study, which uses a fixed backbone to analyse cross-sensor generalisation, and a representation study, which compares different backbone and initialisation choices.
Across the three tasks, we evaluate (A) within-sensor training and  (B) zero-shot cross-sensor transfer under task-specific protocols,  while (C) few-shot target adaptation is additionally evaluated only for force regression. 
In the within-sensor setting, training and testing 
are performed on the same sensor; in the zero-shot setting, a model trained  on a source sensor is evaluated directly on a target sensor without using  any target labels. 
For shape classification, we conduct an exhaustive 
$7 \times 7$ source-target evaluation, in which each of the seven sensors is used in turn as both the source and the target. 
For grating classification and force regression, we adopt controlled fixed-source one-to-many evaluation protocols: GelSightMarker and GelSightNoMarker are fixed as the source sensors  for the two tasks, respectively, while the target sensor is varied. 
Target-specific within-sensor models are also trained and evaluated as reference baselines. 
Few-shot target adaptation is evaluated only for force regression. Starting from the source-trained force-regression model, both the ViT backbone and the regression head are fine-tuned using a small labelled subset from the target-sensor training partition. 
For each target sensor, $0.5\%$, $1\%$, $2.5\%$, $5\%$, and $10\%$ of the training samples are 
randomly selected without replacement. 
The labelled target-sensor validation set is used for model selection and early stopping, while the target-sensor test set remains untouched.


Before dataset partitioning, temporally adjacent frames were compared using the structural similarity index measure (SSIM). 
A frame was treated as redundant and removed when its SSIM relative to the most recently retained frame was greater than or equal to 0.99. 
This conservative threshold was used to remove near-identical observations while preserving subtle tactile variations.
This filtering procedure was completed before dataset partitioning to reduce the possibility that highly similar temporal neighbours were assigned to different subsets.

For shape classification, retained samples belonging to the same contact trial were first grouped together. The contact trials were then stratified by sensor and shape class, ordered according to acquisition time, and divided into contiguous 60/20/20 blocks for training, validation, and testing, respectively. Thus, no contact trial appears in more than one partition.
For grating classification and force regression, the complete trial or trajectory was used as the partitioning unit. 
All frames belonging to the same trial or trajectory were assigned exclusively to one subset.
The ordered trials or trajectories were divided chronologically, with the first 60\% used for training, the following 20\% for validation, and
the remaining 20\% for testing. 
Consequently, no trial or trajectory appears in more than one partition. 
Few-shot target samples are drawn exclusively from the target-sensor training partition, while the
validation and test partitions remain unchanged.

We conducted experiments on a single server equipped with four NVIDIA H200 GPUs. For the transfer study, we use a single ViT backbone with ImageNet initialisation so that performance differences primarily reflect sensor transfer rather than architectural variation. All input images are resized to $224 \times 224$. Downstream models are trained for up to 30 epochs using AdamW with batch size 128, learning rate $1\times10^{-4}$, weight decay $1\times10^{-4}$, cosine learning-rate decay with warmup, gradient clipping at 1.0, and early stopping with patience 10. All runs use a fixed random seed of 42.
For the representation study, we compare ResNet-18~\cite{he2016deep}, ViT trained from scratch, ViT with ImageNet pretraining~\cite{dosovitskiy2020image}, 
and ViT initialised from Masked Autoencoder (MAE) pretraining~\cite{he2022masked}. 
The objective of the transfer study is to quantify the effect of sensor shift under an architecture-controlled setting.

The MAE framework is used to learn tactile representations without using task annotations. Given an input tactile image, MAE divides it into non-overlapping patches and
randomly masks a proportion $r$ of them. 
A ViT encoder processes only the visible patches, while a lightweight decoder combines
the encoded features with learnable mask tokens to reconstruct the missing patches.

The encoder and decoder are jointly trained by minimising the reconstruction error over the masked patches:
$
\mathcal{L}_{\mathrm{MAE}}
=
\frac{1}{|\mathcal{M}|}
\sum_{i\in\mathcal{M}}
\left\|
    \widehat{\mathbf{x}}_{i}
    -
    \mathbf{x}_{i}
\right\|_{2}^{2},
$
where $\mathcal{M}$ denotes the set of masked patches, and
$\mathbf{x}_{i}$ and $\widehat{\mathbf{x}}_{i}$ denote the original and
reconstructed pixel values of patch $i$, respectively. By inferring the
missing content from visible regions, the encoder learns spatial
dependencies among contact geometry, deformation patterns, and
sensor-specific visual cues.

For TacVerse, the shared tactile encoder is pretrained in a self-supervised manner using only the unlabelled tactile images from the training partitions of the shape classification, grating classification, and force regression tasks, while all validation and test images are excluded from the pretraining corpus.
After pretraining, the decoder is discarded, and the pretrained encoder is transferred to task-specific classification or regression heads for downstream
fine-tuning. 
Pretraining is performed for 200 epochs using AdamW with a batch size of 256, an initial learning rate of $1.5\times10^{-4}$,
weight decay of $0.05$, and a masking ratio of
$r=$ 0.5.

For the classification tasks, we report accuracy, precision, recall, and F1. 
Accuracy denotes the proportion of correctly classified samples. 
Precision, recall, and F1 are computed independently for each class using a one-vs-rest formulation and then averaged equally across all classes. 
For force regression, we report MAE, MSE, RMSE, and $R^2$.


\subsection{Shape Classification}

\begin{figure}[!t]
\centering
\includegraphics[width=1\columnwidth]{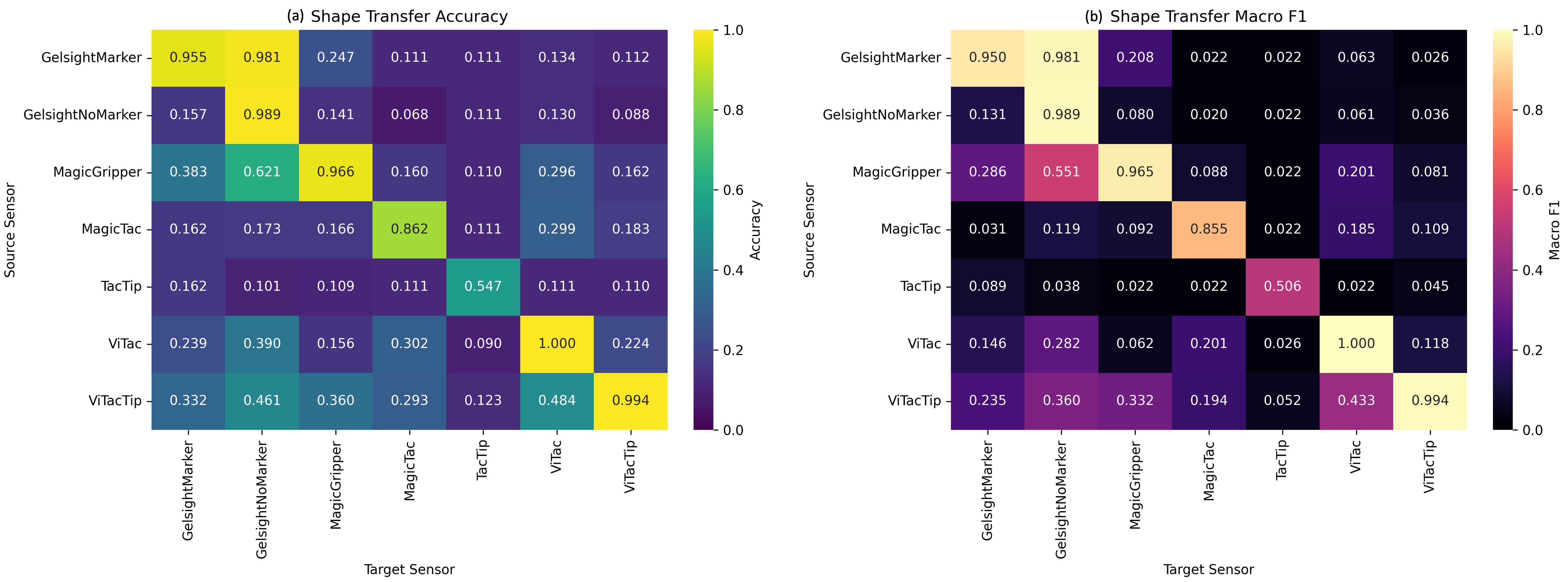}
\caption{Heatmap of shape-classification transfer performance across source and target sensors. The diagonal entries indicate within-sensor results, while off-diagonal entries reveal substantial sensor-shift effects; transfer is strongest between the two GelSight variants and remains limited for more dissimilar sensor pairs.}
\label{exp-1}
\end{figure}

For shape classification, we conduct source-target evaluation across all seven sensors, with the complete results reported as accuracy and F1 heatmaps in Figure~\ref{exp-1}. 
To provide detailed precision, recall, and F1 results, Table~\ref{table:shape} reports three representative source sensors: TacTip (MDM), GelSightMarker (IMM+MDM), and MagicTac (IMM+MDM+MFM). These three sources span single-, two-, and three-modality sensing configurations, respectively. 
Overall, the results show that within-sensor training consistently provides the best performance, while cross-sensor transfer remains highly dependent on the similarity between source and target sensors.

\begin{table*}[!t]
\centering
\caption{Shape-classification transfer results for TacTip, GelSightMarker, and MagicTac as source sensors. 
The best result is \textbf{bolded} and the second best is \underline{underlined}.}
\begin{tabular}{l|l|c|cccc}
\toprule
\textbf{Source Sensor} & \textbf{Target Sensor} & \textbf{Protocol} & \textbf{Accuracy$\uparrow$} & \textbf{Precision$\uparrow$} & \textbf{Recall$\uparrow$} & \textbf{F1$\uparrow$} \\
\midrule
\multirow{7}{*}{TacTip}
& TacTip           & within-sensor & \textbf{0.547} & \textbf{0.636} & \textbf{0.547} & \textbf{0.506} \\
& GelSightMarker   & cross-sensor & \underline{0.162} & \underline{0.224} & \underline{0.124} & \underline{0.089} \\
& GelSightNoMarker & cross-sensor & 0.101 & 0.029 & 0.101 & 0.038 \\
& MagicGripper     & cross-sensor & 0.109 & 0.012 & 0.109 & 0.022 \\
& MagicTac         & cross-sensor & 0.111 & 0.012 & 0.111 & 0.022 \\
& ViTac            & cross-sensor & 0.111 & 0.012 & 0.111 & 0.022 \\
& ViTacTip         & cross-sensor & 0.110 & 0.031 & 0.110 & 0.045 \\
\midrule
\multirow{7}{*}{GelSightMarker}
& GelSightMarker   & within-sensor & \underline{0.955} & \underline{0.969} & \underline{0.936} & \underline{0.950} \\
& GelSightNoMarker & cross-sensor  & \textbf{0.981} & \textbf{0.982} & \textbf{0.981} & \textbf{0.981} \\
& MagicGripper     & cross-sensor  & 0.247 & 0.629 & 0.247 & 0.208 \\
& MagicTac         & cross-sensor  & 0.111 & 0.012 & 0.111 & 0.022 \\
& TacTip           & cross-sensor  & 0.111 & 0.012 & 0.111 & 0.022 \\
& ViTac            & cross-sensor  & 0.134 & 0.146 & 0.134 & 0.063 \\
& ViTacTip         & cross-sensor  & 0.112 & 0.124 & 0.112 & 0.026 \\
\midrule
\multirow{7}{*}{MagicTac}
& MagicTac         & within-sensor & \textbf{0.862} & \textbf{0.877} & \textbf{0.862} & \textbf{0.855} \\
& GelSightMarker   & cross-sensor & 0.162 & 0.018 & 0.111 & 0.031 \\
& GelSightNoMarker & cross-sensor & 0.173 & 0.118 & 0.173 & 0.119 \\
& MagicGripper     & cross-sensor & 0.166 & \underline{0.175} & 0.166 & 0.092 \\
& TacTip           & cross-sensor & 0.111 & 0.012 & 0.111 & 0.022 \\
& ViTac            & cross-sensor & \underline{0.299} & 0.162 & \underline{0.299} & \underline{0.185} \\
& ViTacTip         & cross-sensor & 0.183 & 0.117 & 0.183 & 0.109 \\
\bottomrule
\end{tabular}
\label{table:shape}
\end{table*}

\begin{figure}[!t]
\centering
\includegraphics[width=1.0\columnwidth]{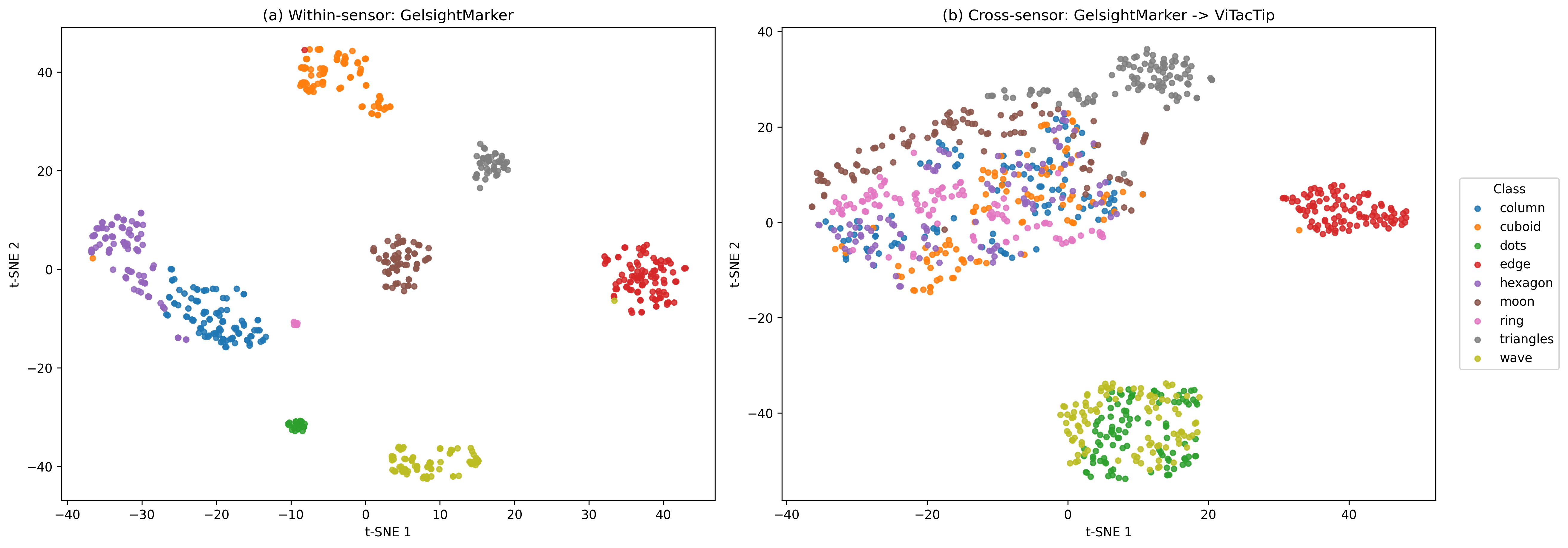}
\vspace{-0.4cm}
\caption{t-SNE visualisation of shape-task features under within-sensor and cross-sensor settings. Within-sensor embeddings form tighter, better-separated class clusters, whereas cross-sensor embeddings become more dispersed and overlapping, highlighting the representation shift caused by sensor mismatch.}

\label{exp-2}
\end{figure}

Figure~\ref{exp-1} summarises within-sensor and cross-sensor performance across seven sensors using accuracy and F1 heatmaps. As expected, the diagonal entries, corresponding to the within-sensor setting, achieve the best results for all evaluated source sensors. A particularly interesting finding is that the model trained on GelSightMarker achieves even higher zero-shot accuracy on GelSightNoMarker than on its original test domain. This suggests that the strong structural similarity between the two GelSight variants can outweigh the effect of domain shift and, in some cases, even improve generalisation. By contrast, transfer to more dissimilar sensors degrades markedly, indicating that shape representations are still sensitive to differences in sensing mechanism and image formation.

Figure~\ref{exp-2} provides a t-SNE visualisation of the learned feature space. The embeddings show clearer and more compact clustering in the within-sensor setting, whereas cross-sensor samples become more scattered and overlapping. This trend is consistent with the quantitative results and further illustrates that sensor mismatch leads to a noticeable shift in representation space, even for a comparatively robust task such as shape classification.

\subsection{Grating Classification}
\begin{table}[!t]
\centering 
\caption{Grating classification results with GelSightMarker fixed as the source sensor. Performance is the best under within-sensor evaluation and drops markedly under sensor shift; among the cross-sensor targets, GelSightNoMarker remains the most transferable, while MagicGripper and ViTacTip show near-chance performance.}
\label{tab:grating}
\begin{tabular}{l|l|c|cccc}
\toprule
\textbf{Source Sensor} & \textbf{Target Sensor} & \textbf{Protocol} & \textbf{Accuracy$\uparrow$} & \textbf{Precision$\uparrow$} & \textbf{Recall$\uparrow$} & \textbf{F1$\uparrow$} \\
\midrule
\multirow{4}{*}{GelSightMarker}
& GelSightMarker   & within-sensor & \textbf{0.903} & \textbf{0.906} & \textbf{0.903} & \textbf{0.903} \\
& GelSightNoMarker & cross-sensor  & \underline{0.248} & \underline{0.315} & \underline{0.248} & \underline{0.238} \\
& MagicGripper     & cross-sensor  & 0.054 & 0.023 & 0.054 & 0.013 \\
& ViTacTip         & cross-sensor  & 0.041 & 0.005 & 0.041 & 0.006 \\
\bottomrule
\end{tabular}
\end{table}

For grating classification, we adopt a controlled fixed-source one-to-many protocol, using GelSightMarker as the source sensor. 
The source training domain is held constant while the target sensor is varied, allowing performance differences to be attributed primarily to target-sensor shift rather than to changes in the source training distribution. 
Table~\ref{tab:grating} reports the within-sensor reference and the corresponding cross-sensor results under this controlled setting.
The within-sensor setting yields the best performance, with accuracy, precision, recall, and F1 all reaching 0.903 or above on GelSightMarker. This confirms that the grating patterns are highly learnable when the training and test data come from the same sensor domain. Under cross-sensor evaluation, however, performance degrades substantially. Although GelSightNoMarker remains the most transferable target, its accuracy drops from 0.903 to 0.248, which is notably worse than the corresponding transfer behaviour observed in the shape-classification task. Transfer to MagicGripper and ViTacTip is even weaker, with results close to chance level, indicating that grating recognition is highly sensitive to sensor-specific image formation and texture appearance.

\begin{figure}[!t]
\centering
\includegraphics[width=0.9\columnwidth]{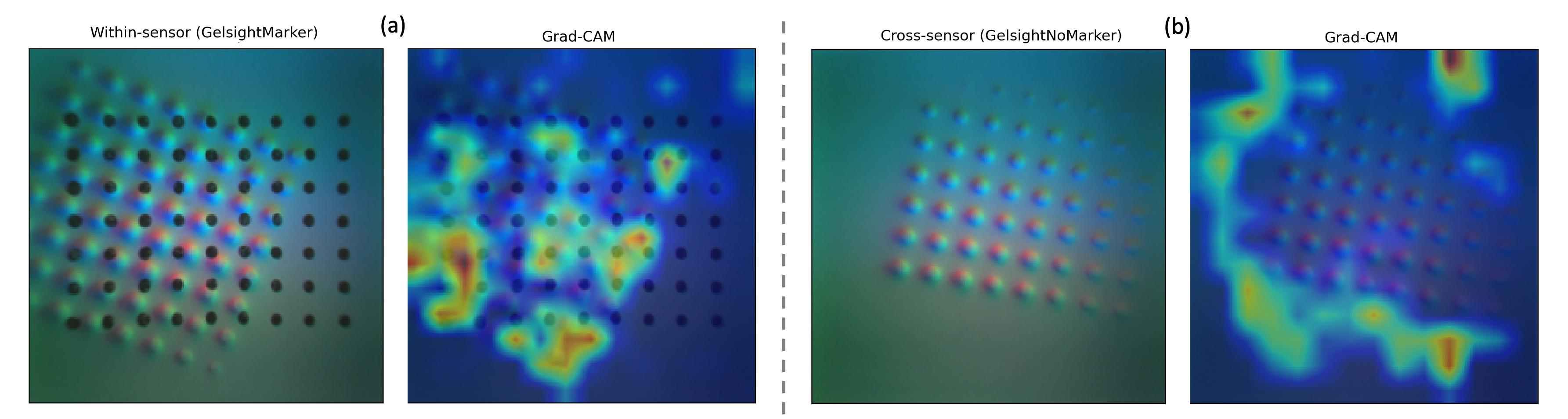}
\caption{Grad-CAM visualisations for grating classification under within-sensor and cross-sensor evaluation using the same model trained on GelSightMarker. For samples from the same sensor domain as the training data, the model attends to the contact region and captures the grating pattern effectively. Under cross-sensor transfer, the attention becomes more diffuse and tends to surround rather than focus on the target region, indicating that sensor mismatch weakens localisation of the most discriminative cues.}
\label{exp-3}
\end{figure}

Figure~\ref{exp-3} provides qualitative evidence for this degradation through Grad-CAM visualisations. In the within-sensor setting, the model focuses clearly on the contact region and captures the discriminative grating structure. In the cross-sensor setting, by contrast, the activation maps become more diffuse and tend to surround rather than concentrate on the target area. This suggests that sensor mismatch not only reduces classification accuracy but also weakens the model's ability to localise the most informative tactile cues.

\subsection{Force Regression}
For force regression, we similarly adopt a controlled fixed-source one-to-many protocol, with GelSightNoMarker serving as the source sensor. 
The model is trained once on this fixed source domain and evaluated on multiple target sensors. This design isolates the effect of target-sensor shift and enables direct comparison of force regression performance across target domains. 
As shown in Table~\ref{tab:force},
the within-sensor setting again produces the strongest performance, achieving the lowest MAE, MSE, and RMSE, together with the highest $R^2$ value. Under cross-sensor evaluation, however, performance deteriorates substantially. Although GelSightMarker remains the most transferable target among the three cross-sensor settings, its RMSE still increases from 0.186 to 0.348 and its $R^2$ drops below zero, indicating that the model no longer provides reliable force estimation after sensor shift. Transfer to MagicGripper and ViTacTip is markedly worse, with much larger prediction errors and strongly negative $R^2$ values, showing that force regression is particularly sensitive to changes in sensor structure and contact-response characteristics.

\begin{figure}[!t]
\centering
\includegraphics[width=1\columnwidth]{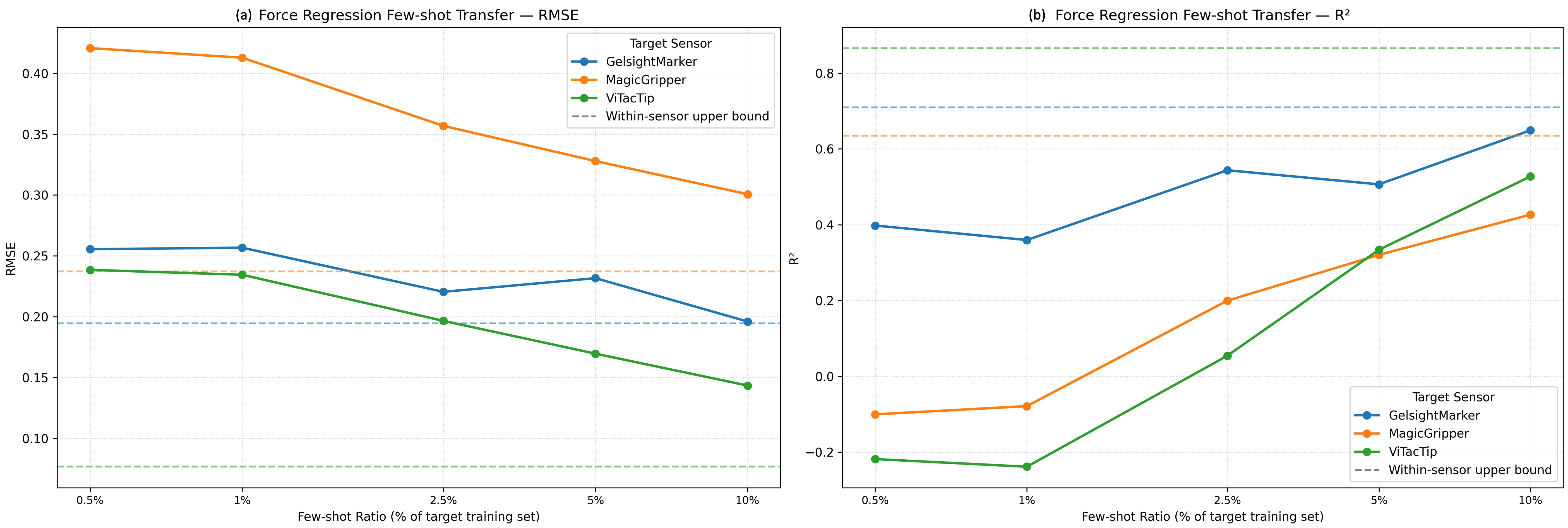}
\vspace{-0.5cm}
\caption{Few-shot adaptation results for force regression on three target sensors. As the proportion of labelled target data increases, RMSE decreases and $R^2$ improves consistently, but the adapted models remain below the dashed within-sensor upper bounds, indicating that few-shot supervision narrows but does not remove the cross-sensor performance gap.}
\label{exp-4}
\end{figure}

Figure~\ref{exp-4} further illustrates the effect of few-shot adaptation for three target sensors. As the few-shot ratio increases, RMSE decreases progressively and $R^2$ rises accordingly, confirming that even a small amount of target-domain supervision helps the model recover useful force-related representations. Nevertheless, the adapted models do not surpass the dashed within-sensor upper bounds for their corresponding target sensors. This indicates that few-shot learning can partially mitigate the cross-sensor gap in force estimation, but does not fully eliminate the advantage of training and testing on the same sensor.

\begin{table*}[!t]
\centering
\caption{Force regression results with GelSightNoMarker fixed as the
source sensor. Evaluation on GelSightNoMarker is within-sensor, whereas evaluation on the remaining target sensors is zero-shot cross-sensor transfer.}
\label{tab:force}
\begin{tabular}{l|l|c|cccc}
\toprule
\textbf{Source Sensor} & \textbf{Target Sensor} & \textbf{Protocol} & \textbf{MAE$\downarrow$} & \textbf{MSE$\downarrow$} & \textbf{RMSE$\downarrow$} & \textbf{R$^2$$\uparrow$} \\
\midrule
\multirow{4}{*}{GelSightNoMarker}
& GelSightNoMarker & within-sensor     & \textbf{0.126} & \textbf{0.035} & \textbf{0.186} & \textbf{0.590} \\
& GelSightMarker   & cross-sensor & \underline{0.287} & \underline{0.126} & \underline{0.348} & \underline{-0.014} \\
& MagicGripper     & cross-sensor & 0.766 & 0.954 & 0.976 & -6.046 \\
& ViTacTip         & cross-sensor & 0.863 & 1.631 & 1.277 & -9.058 \\

\bottomrule
\end{tabular}
\end{table*}

\begin{table}[!t]
\centering
\caption{Representation study across backbones and initialisation strategies on three tasks.}
\label{tab:representation_study}
\begin{tabular}{cl cccc}
\toprule
Task & Sensor & ResNet-18  & ViT\textsubscript{Rand} & ViT\textsubscript{ImgNet} & ViT\textsubscript{MAE} \\
\midrule
\multirow{4}{*}{\makecell[c]{Shape\\(Acc$\uparrow$)}}
 & GelSightMarker   &  \textbf{0.966}&  0.422&  \underline{0.958}&  0.848\\
 & GelSightNoMarker &  \underline{0.984}&  0.706&  \textbf{0.989}&  0.942\\
 & MagicGripper     &  \underline{0.931}&  0.716&  0.908&  \textbf{0.949}\\
 & ViTacTip         &  \underline{0.992}&  0.877&  0.991&  \textbf{0.996}\\
 \midrule
 \multirow{4}{*}{\makecell[c]{Grating\\(Acc$\uparrow$)}}
 & GelSightMarker   &  0.808&  0.552&  \underline{0.887}&  \textbf{0.944}\\
 & GelSightNoMarker &  0.711&  0.576&  \underline{0.768}&  \textbf{0.938}\\
 & MagicGripper     &  0.595&  0.393&  \underline{0.698}&  \textbf{0.757}\\
 & ViTacTip         &  0.757&  0.116&  \textbf{0.817}&  0.673\\
\midrule
\multirow{4}{*}{\makecell[c]{Force\\(RMSE$\downarrow$)}}
 & GelSightMarker   &  0.241&  0.375&  \underline{0.197}&  \textbf{0.158}\\
 & GelSightNoMarker &  0.258&  0.335&  \underline{0.186}&  \textbf{0.170}\\
 & MagicGripper     &  0.290&  0.332&  \underline{0.266}&  \textbf{0.240}\\
 & ViTacTip         &  0.063&  0.190&  \underline{0.059}&  \textbf{0.057}\\
 \bottomrule
\end{tabular}
\end{table}

\subsection{Representation Study}
Finally, we compare backbone and initialisation choices under within-sensor training to assess whether self-supervised pretraining on TacVerse can provide a unified representation across tasks. Table~\ref{tab:representation_study} reports results for four sensors shared across the shape, grating, and force benchmarks. 
Overall, the representation study indicates that model initialisation has a substantial effect on downstream performance, with MAE pretraining emerging as the most effective choice across most task–sensor combinations.

For shape classification, the best results are distributed across ImageNet-pretrained ViT and MAE-pretrained ViT, with both clearly outperforming the randomly initialised ViT. MAE achieves the top accuracy on MagicGripper and ViTacTip, while ImageNet pretraining performs best on GelSightNoMarker and remains highly competitive on the other sensors. This suggests that shape recognition benefits from both generic visual priors and tactile-specific pretraining, especially when the sensor observations preserve stable geometric structure.

For grating classification, the advantage of MAE pretraining becomes more pronounced. The MAE-initialised ViT achieves the best accuracy on GelSightMarker, GelSightNoMarker, and MagicGripper, substantially outperforming both the randomly initialised ViT and ResNet-18. Only on ViTacTip does the ImageNet-pretrained ViT achieve the best result. These trends indicate that fine-grained grating recognition depends strongly on high-quality feature initialisation, and that self-supervised tactile pretraining is especially effective for capturing subtle texture-related cues.

For force regression, MAE pretraining yields the lowest RMSE on all four sensors, again demonstrating the strongest overall transferability across tasks. Although the ImageNet-pretrained ViT is consistently the second-best performer, the margin in favour of MAE remains clear, particularly on GelSightMarker and MagicGripper. Taken together, these results show that MAE pretraining on TacVerse produces a robust and versatile tactile representation that generalises well across classification and regression settings, 
supporting its use as a shared self-supervised initialisation for
vision-based tactile perception.

\section{Conclusion and Discussion}
We present TacVerse, a multi-sensor benchmark containing 106,800 tactile images from seven vision-based tactile sensors and covering shape classification, grating classification, and force regression. Its distinctive value lies in the use of shared task definitions and unified within-sensor, direct cross-sensor, and few-shot adaptation protocols, which enable controlled measurement of sensor shift. Our results show that strong within-sensor performance does not necessarily translate across sensor platforms. Shape classification is comparatively robust, whereas grating classification and force regression are more sensitive to sensor mismatch. Few-shot adaptation for force regression reduces this performance gap, and tactile MAE pretraining provides the most consistent improvements across the evaluated tasks.

TacVerse currently focuses on controlled laboratory interactions, while the grating and force benchmarks adopt fixed-source evaluation. 
The current study also does not provide a comprehensive comparison of specialised tactile foundation models. 
Future work can expand TacVerse toward broader sensor and task coverage, dynamic tactile interactions, and real-world manipulation. 
Another important direction is to investigate sensor-invariant representation learning, more data-efficient adaptation, and recent tactile foundation models under a unified evaluation protocol.

\section{Code and Data Availability}
To support reproducibility and future research on cross-sensor tactile perception, we release the resources associated with TacVerse through the following channels. 
The project code, training scripts, and experiment configurations are
available on GitHub at \url{https://github.com/LannWei/Tactile_Database}.
The TacVerse dataset is released on Hugging Face at \url{https://huggingface.co/datasets/Lan-2025/Tactile}. 
Project website is available at \url{https://lannwei.github.io/Tactile_Database/}.


\medskip
\bibliographystyle{MSP}
\bibliography{root}

\end{document}